\documentclass[preprint]{article}

\usepackage{neurips_2023}

\usepackage{bookmark}
\usepackage[utf8]{inputenc} 
\usepackage[T1]{fontenc}    
\usepackage{hyperref}       
\usepackage{url}            
\usepackage{booktabs}       
\usepackage{amsfonts}       
\usepackage{nicefrac}       
\usepackage{microtype}      
\usepackage{xcolor}         
\usepackage{listings}
\usepackage{cite}
\usepackage{graphicx}
\usepackage{hyperref}
\usepackage{amsmath}
\usepackage{algpseudocode}
\usepackage{amsmath}
\usepackage{algorithm}
\usepackage{caption}
\usepackage{pdflscape}
\usepackage{tabularx}

\hypersetup{
    pdftitle={Adapting Safe-for-Work Classifier for Malaysian Language Text: Enhancing Alignment in LLM-Ops Framework},
    pdfauthor={Aisyah Razak, Ariff Nazhan, Ariff Nazhan},
    pdfsubject={Natural Language Processing, Large Language Models, Machine Learning},
    pdfkeywords={language models, malay natural language processing, deep learning, text classification},
    colorlinks=true,
    linkcolor=blue,
    citecolor=blue,
    urlcolor=blue
}
\title{Adapting Safe-for-Work Classifier for Malaysian Language Text: Enhancing Alignment in LLM-Ops Framework}
\author{
  Aisyah Razak\thanks{aisyahrazak171@gmail.com} \and
  Ariff Nazhan\thanks{ariffnzhn@gmail.com} \and
  Kamarul Adha\thanks{kamarul.adha360@gmail.com} \and
  Wan Adzhar Faiq Adzlan\thanks{adzhar.faiq@gmail.com} \and
  Mas Aisyah Ahmad\thanks{masaisyahahmad@gmail.com} \and
  Ammar Azman\thanks{ammarakef98@gmail.com} \and
}

\begin{document}

\maketitle

\begin{abstract}

  As large language models (LLMs) become increasingly integrated into operational workflows (LLM-Ops), there is a pressing need for effective guardrails to ensure safe and aligned interactions, including the ability to detect potentially unsafe or inappropriate content across languages. However, existing safe-for-work classifiers are primarily focused on English text. To address this gap for the Malaysian language, we present a novel safe-for-work text classifier tailored specifically for Malaysian language content. By curating and annotating a first-of-its-kind dataset of Malaysian text spanning multiple content categories, we trained a classification model capable of identifying potentially unsafe material using state-of-the-art natural language processing techniques. This work represents an important step in enabling safer interactions and content filtering to mitigate potential risks and ensure responsible deployment of LLMs. To maximize accessibility and promote further research towards enhancing alignment in LLM-Ops for the Malaysian context, the model is publicly released at \href{https://huggingface.co/malaysia-ai/malaysian-sfw-classifier}{malaysia-ai/malaysian-sfw-classifier}.
\end{abstract}

\section{Introduction}

The AI field, especially natural language processing, \cite{markov2023holistic} has seen remarkable progress with significant breakthroughs like transformer-based architectures \cite{vaswani2023attentionneed}, multimodality integration to chatbot applications, and reinforcement learning from human feedback. This has led to the rise of open-domain dialogue systems, known as chatbots or conversational agents, which are now increasingly integrated into our daily lives.

Due to the nature of how large language models are trained, using internet data, it is prevalent that there may be harmful contents included. However, as users continues engaging with these chatbots, exposure to harmful and provocative text can have significant adverse effects, impacting individuals' mental well-being, relationships, and emotional state. Therefore, ensuring safe and beneficial interactions has become critically important.

The scarcity of data for identifying not safe for work content, particularly in the Malay language, hinders the advancement of undesired content filtration. Past work such as \cite{markov2023holistic,qiu2024facilitatingpornographictextdetection}  have laid foundation in terms of AI moderation using large language models, but there is still no not safe for work task in malay language. In this paper, we address this challenge by initiating the data gathering process to create a comprehensive dataset of harmful texts. Our methodology involves mining data representative of harmful text categories. Our categorization includes the following labels: pornography, harassment, sexist, racist, religious insult, self-harm, psychiatric or mental illness, and safe for work.

We aim to create a robust classifier tailored for Malaysian language text, enhancing the alignment of our large language model operations framework with safety and ethical standards. This classifier serves as a necessary guardrail within the LLM-Ops framework, providing a cost-effective solution for ensuring safe AI. By systematically identifying and filtering out inappropriate content, this classifier will help create a safe and respectful interaction environment for users.

Furthermore, to the best of our knowledge, there is currently no existing local dataset for the Malaysian language that addresses these specific categories of harmful content. Our work thus represents a pioneering effort in developing and applying this crucial safety measure.

\section{Taxonomy}

Designing a universal taxonomy for safe for work guardrails is challenging due to the context-dependent nature of language. Below, we outline our taxonomy for safe-for-work categorization, which will guide the application of guardrails to our large language model or chatbot system. Each category is described to clarify the scope and specifics of what constitutes undesired content:

\begin{itemize}
  \item \textbf{Pornography:} Content that includes explicit sexual descriptions, depictions of sexual acts, or nudity intended to arouse sexual interest. This category covers sexually explicit text, adult content descriptions, and any language or media that depicts sexual activity.

  \item \textbf{Harassment:} Content that targets individuals or groups with the intent to demean, intimidate, or threaten. This includes abusive language, threats, stalking, or any form of verbal harassment aimed at causing emotional or psychological distress.

  \item \textbf{Sexist:} Content that promotes discrimination or prejudice based on gender. This includes sexist remarks, derogatory comments about any gender, and language that reinforces harmful gender stereotypes or inequality.

  \item \textbf{Racist:} Content that discriminates or promotes hatred based on race, ethnicity, or nationality. This includes racial slurs, derogatory remarks about ethnic groups, and any language that supports racial superiority or inferiority.

  \item \textbf{Religious Insult:} Content that disrespects or mocks religious beliefs, practices, or figures. This includes blasphemy, offensive jokes about religions, and language intended to insult or offend individuals based on their religious affiliations.

  \item \textbf{Self-Harm:} Content that depicts or encourages self-injurious behavior or suicide. This includes descriptions of self-harm methods, discussions promoting suicide, and any language that glorifies or encourages self-destructive actions.

  \item \textbf{Psychiatric or Mental Illness:} Content that stigmatizes or discriminates against individuals with mental health conditions. This includes derogatory terms for mental health issues, insensitive jokes, signs of mental distress or illness, and any language that trivializes or mocks mental illness.

  \item \textbf{Violence:} Content that promotes or glorifies violence or celebrates the suffering or humiliation of others.

  \item \textbf{Safe for Work:} Content that is appropriate for a professional or public environment, free from explicit, offensive, or discriminatory material. This includes clean language, respectful discourse, and content that does not contain any of the above-mentioned undesired elements.
\end{itemize}

This taxonomy will help us systematically identify and filter out inappropriate content, where additional redirection can be made to ensure a safe and respectful interaction environment for users.

\section{Data Source}
The data for this study was collected from various platforms, including social media, public forums, and publicly available datasets. The majority of the data is in the malay language and relevant to the malay context.
By utilizing these comprehensive datasets from multiple sources, we have strengthened the robustness and accuracy of our classification model, enabling it to effectively tackle the challenges of identifying harmful topics in online content.
\subsection{Social Media}
Data was collected from popular social media platforms such as Twitter and Facebook. Two main approaches were employed to gather relevant data:

1. Keyword-based scraping: a list of keywords associated with explicit content was compiled. These keywords were used to extract tweets from the platform.

2. Profile-based scraping: a list of profiles known for regularly posting NSFW content was curated. Posts from these profiles were then scraped to obtain a more targeted dataset.

The combination of these two scraping methods resulted in a comprehensive and diverse dataset from social media, capturing both keyword-specific content and data from profiles that frequently share explicit material.

\subsection{Public Articles}

For public articles, we have collected data from various articles and blogs which are \href{http://b.cari.com.my}{b.cari}, which hosts a wide range of user-generated content in Malay.

The collected dataset consists of human dialogues extracted from these articles and blog posts. It notably includes some dialogues that contain nsfw content. The inclusion of such dialogues, while potentially controversial, is important to allow the trained classifier to effectively detect explicit content that may realistically occur in open-ended dialogue systems.

\subsection{Public Datasets}

We have also collected data from publicly available dataset on kaggle such as \href{https://www.kaggle.com/datasets/nikhileswarkomati/suicide-watch}{Kaggle: Suicide and Depression Detection} datasets. The dataset is a collection of posts from the "SuicideWatch" and "depression" subreddits of the Reddit platform. This dataset contains a wide range of suicide ideation contexts, providing valuable insights for our research.

We also leverage \href{https://github.com/rewire-online/edos}{Explainable Detection of Online Sexism (EDOS)} from github and \href{http://nlp.uned.es/exist2021/}{EXIST: sEXism Identification in Social neTwork (EXIST)} from web that contain a diverse collection of sexism statements, which have significantly contributed to the success of our classifier in identifying and categorizing such content.

\section{Methodology}

Supervised text classification requires reliable class labels for training data. However, obtaining these labels can be complex and expensive. Typically, labels are added sequentially by querying an annotator until satisfactory performance is achieved. We introduced an approach that leverages active learning, knowledge distillation of large language models, and text clustering to reduce annotation effort and construct a collection of labeled not safe for work (NSFW) data from our gathered data.

Following figure illustrates the overall flow employed in our methodology to collect NSFW dataset aims to label malaysian dataset for alignment in LLMOps framework.

\begin{figure}[h]
  \centering
  \includegraphics[width=0.8\linewidth]{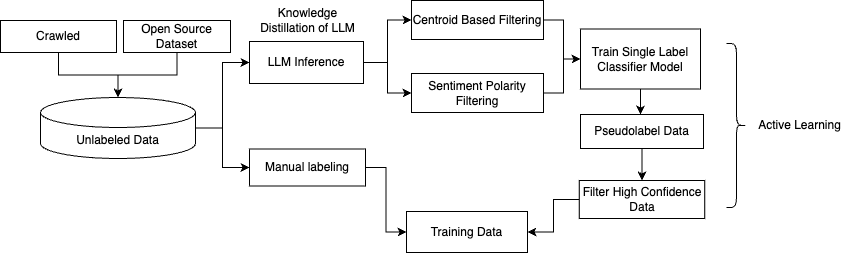}
  \caption{Overall Model Training Framework}
\end{figure}

\subsection{Manual Labeling}
Relying on large language models (LLMs) for data labeling can present challenges, such as inaccuracies and biases introduced by the model. To address these issues and enhance the quality of our dataset, we employed manual labeling using Label Studio \cite{LabelStudio} to complement dataset produced from the label produced by large language models.

Label Studio \cite{LabelStudio} is an open-source data labeling tool that allows for efficient, precise annotation of data and we use it to manually label approximately 200 data points, which were then added to our dataset as a baseline.

\begin{itemize}
  \item \textbf{Annotation Process}: Our team meticulously reviewed and annotated the data points to ensure high-quality labels.
  \item \textbf{Baseline Dataset}: The manually labeled data serves as a baseline for our model. By starting with a small, accurately labeled dataset, we provide the model with a strong foundation for learning.
\end{itemize}

\begin{figure}[h]
  \centering
  \includegraphics[width=0.6\linewidth]{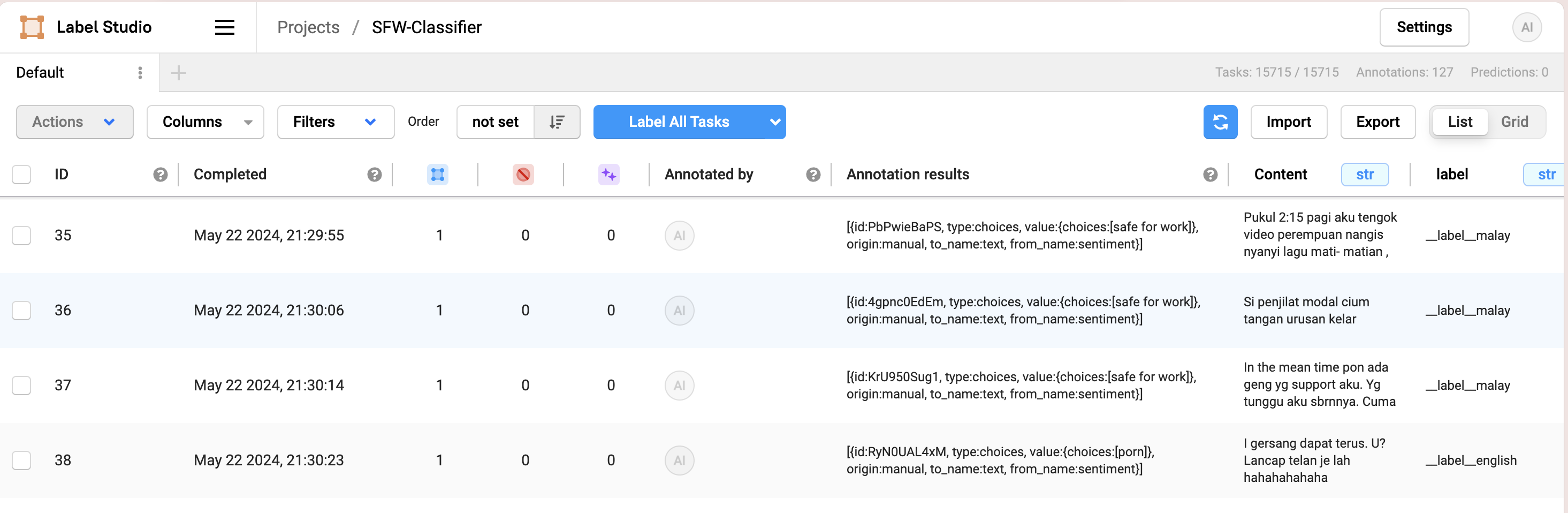}
  \caption{Manual labeling process using Label Studio \cite{LabelStudio}.}
\end{figure}

\subsection{Knowledge Distillation of Large Language Models}

Data annotation using knowledge distillation from a large language model (LLM) involves prompting a teacher LLM to generate a label \( y \) for a given input \( x \) as the foundational knowledge. This paper primarily explores the zero-shot paradigm, where the LLM receives only the input \( x \) and an instruction \( I \) without any labeled examples or demonstrations. This straightforward yet effective method of extracting knowledge from teacher LLMs has been widely applied to various tasks and applications. It only requires an input dataset, which is then processed by the LLM to produce the desired label \( y \).

In this annotation stage, we utilized two different large language models: mistral-7b and MaLLaM-small. By leveraging mistral-7b, which excels with English-centric prompts, we translated our scraped data into English to provide better context and achieve reliable outputs from the model. While for MaLLaM \cite{zolkepli2024mallam}, we provided inputs from the original malay text as it has better contextual meaning compared to the other models. Below is the prompt we used to generate the labels from the models:

\begin{lstlisting}[breaklines=true]

    text: <text content>

    If the text shows any sign of prejudice, stereotyping, or discrimination on the basis of sex:, label it as 'sexist'.
    If the text shows any sign of content that threatens, incites, glorifies, or expresses desire for violence or harm., label it as 'violence'.
    If the text shows any sign of prejudice, discrimination, or antagonism by an individual, community, or institution against a person or people on the basis of their membership of a particular racial or ethnic group, label it as 'racist'.
    If the text shows any sign of content depicting adult nudity or sexual behavior that is pornographic or intended to cause sexual arousal, label it as 'porn'.
    If the text shows any sign of psychiatric or mental illness, label it as 'psychiatric or mental illness'.
    If the text shows any sign of promotion, or otherwise encourage, suicide or self-harm, label it as 'self-harm'.
    If the text shows any sign of harassment, label it as 'harassment'.
    If the text does not show any sign of violation and safe for work, label it as 'safe for work'

    Only use the label from above choice.

    return the result in JSON format {'label', 'explain'}

\end{lstlisting}

\subsection{Centroid Based Filtering}

We utilize labeled data obtained from a large language model (LLM) to enhance the quality and consistency of our dataset. First, we compute the centroid of the feature vectors for the labeled data, representing the central point in the feature space. The determination of similarity is guided by the Euclidean distance formula, where lower values indicate greater similarity and higher values signify greater dissimilarity.

\begin{algorithm}[H]
  \caption{Centroid-Based Filtering}
  \label{alg:centroid_filtering}
  \begin{algorithmic}[1]
    \State \textbf{Input:} Labeled text data, LLM for embeddings
    \State \textbf{Output:} Filtered dataset
    \State Generate embeddings for the specific topic labeled by LLM
    \State Compute the centroid of the embeddings
    \For{\textbf{each} embedding}
    \State Calculate the Euclidean distance to the centroid
    \EndFor
    \State Observe the distribution of distances
    \State Set a threshold based on the distance distribution
    \State Filter out texts with distances more than the threshold
    \State \textbf{return} Filtered dataset
  \end{algorithmic}
\end{algorithm}

By measuring the distance of each data point from this centroid, we can identify and filter out data points that are far from the centroid. These distant points are likely to be outliers or less representative of the core topic data distribution. This ensures that the labels encapsulate the nuanced semantics of the specific topic. This filtering process enhances the dataset by retaining data that is more coherent and relevant, thereby improving the performance of the classifier.

\begin{figure}[h]
  \centering
  \includegraphics[width=0.6\linewidth]{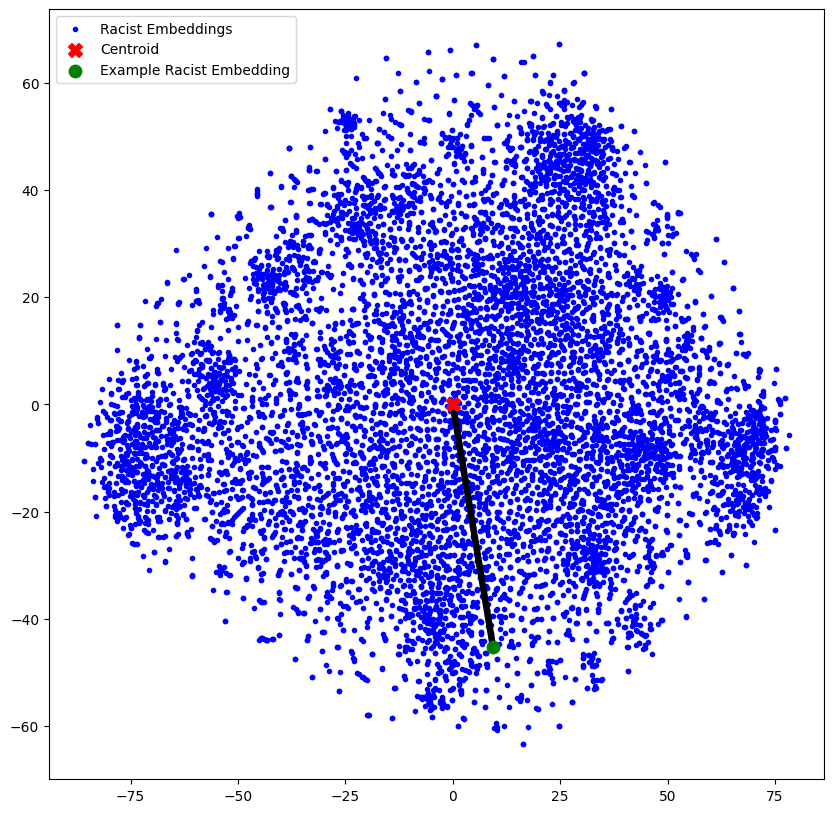}
\end{figure}

\subsection{Sentiment Polarity Filtering}

Ensuring data label accuracy is crucial for optimal model performance, particularly when dealing with the inherently subjective nature of textual data. To improve the quality of our labeled data, we employed sentiment polarity filtering. This approach involves filtering out data labeled with positive sentiment from our dataset, which was initially produced through knowledge distillation of large language models.

For instance, negative sentences like \texttt{Padan muka hang. Tau takut}, categorized under harassment, clearly exhibit negative polarity. In contrast, sentences deemed safe for work typically exhibit positive or neutral polarity. By filtering the dataset based on sentiment polarity, we aim to enhance the dataset's quality, ensuring it includes only the most relevant data.

\subsection{Active Learning}

We use relevant data from a specific category to train a classifier model. The trained model is then used to classify unlabeled data, a process known as pseudolabeling. The outputs given by the classifier are reviewed before feeding all the unlabeled data into a larger classifier model. This iterative process continues until the desired accuracy is achieved.

The following algorithm outlines the active learning process:

\begin{algorithm}[H]
  \caption{Active Learning for Single Label Classifier}
  \begin{algorithmic}[1]
    \State \textbf{Input:} Labeled data \(D_L\), Unlabeled data \(D_U\), Classifier model \(M\), Desired accuracy \(A_{desired}\)
    \State \textbf{Output:} Trained classifier model \(M\)
    \While {Accuracy \(A\) < \(A_{desired}\)}
    \State Train classifier \(M\) on \(D_L\)
    \State Predict labels for \(D_U\) using \(M\) (pseudolabeling)
    \State Filter high-confidence predictions from \(D_U\) to create a new labeled dataset \(D_{L_{new}}\)
    \State Update labeled dataset: \(D_L \leftarrow D_L \cup D_{L_{new}}\)
    \State Retrain classifier \(M\) on the updated \(D_L\)
    \State Manually evaluate the accuracy \(A\) of \(M\)
    \EndWhile
    \State \textbf{return} Trained classifier model \(M\)
  \end{algorithmic}
\end{algorithm}

\section{Result}

In this section, we evaluate the performance of our fine-tuned models using the training set, which is an 80\% split from the NSFW dataset. The remaining 20\% is used to evaluate the performance on the test set. The evaluation metrics used include accuracy, precision, recall, and F1 score. These metrics provide a comprehensive view of the model's performance across different aspects.

\subsection{Model Comparison}

\begin{table}[h]
  \centering
  \caption{Evaluation on Malaysian NSFW Dataset}
  \label{tab:model_comparison}
  \begin{tabular}{lcccccc}
    \hline
    \textbf{Model}                         & \textbf{Accuracy} & \textbf{Precision} & \textbf{Recall} & \textbf{F1 Score} \\
    \hline
    mesolitica/malaysian-mistral-191M-MLM  & 0.8768            & 0.8601             & 0.8854          & 0.8714            \\
    mesolitica/malaysian-mistral-191M-4096 & 0.82583           & 0.81867            & 0.81657         & 0.81556           \\
    microsoft/debertav3-base               & 0.26646           & 0.02961            & 0.11111         & 0.04676           \\
    \hline
  \end{tabular}
\end{table}

Table 1 summarizes the performance metrics for different models. Each model was trained and evaluated with the same test set under the same conditions to ensure a fair comparison.

The results indicate that the \texttt{mesolitica/malaysian-mistral-191M-MLM} model outperforms the other models across all metrics. It achieves the highest accuracy of 0.8768, as well as strong precision, recall, and F1 score values of 0.8601, 0.8854, and 0.8714, respectively. This suggests that \texttt{mesolitica/malaysian-mistral-191M-MLM} is the most effective model for identifying NSFW content in the Malaysian dataset.

The success of the \texttt{mesolitica/malaysian-mistral-191M-MLM} model can be attributed to the implementation of the LLM2Vec \cite{behnamghader2024llm2veclargelanguagemodels} approach. LLM2Vec: Large Language Models Are Secretly Powerful Text Encoders, is a simple and efficient solution to transform any decoder-only LLM into a powerful text encoder in an unsupervised fashion using adapters (LoRA), without the need to modify the base models.

The \texttt{microsoft/debertav3-base} model has the lowest performance among the compared models, with an accuracy of 0.26646 and significantly lower precision, recall, and F1 score values of 0.02961, 0.11111, and 0.04676, respectively. This model is not suitable for NSFW content detection in the Malaysian dataset.

In summary, the \texttt{mesolitica/malaysian-mistral-191M-MLM} model is the most suitable choice for NSFW content detection in the Malaysian context, providing the highest accuracy and consistency across various performance metrics. The \texttt{mesolitica/malaysian-mistral-191M-4096} model, while less effective, still maintains a respectable level of performance. The \texttt{microsoft/debertav3-base} model, however, does not perform adequately for this task.

\subsection{Analysis}

\subsubsection*{2D Embedding}

We convert the texts into embedding representations with a size of 1024 using the \href{https://mesolitica.com/retrieval}{Mesolitica Retrieval API}, which is the current state-of-the-art Malaysian embedding API. After obtaining the embeddings, we decompose them into a 2-dimensional representation using UMAP (Uniform Manifold Approximation and Projection) \cite{2018arXivUMAP} from the UMAP library \cite{mcinnes2018umap-software}.

We sampled 1200 rows for each label to prevent over-scattering and used the default parameters from UMAP for the dimensionality reduction.

\begin{figure}[h]
  \centering
  \includegraphics[width=0.7\linewidth]{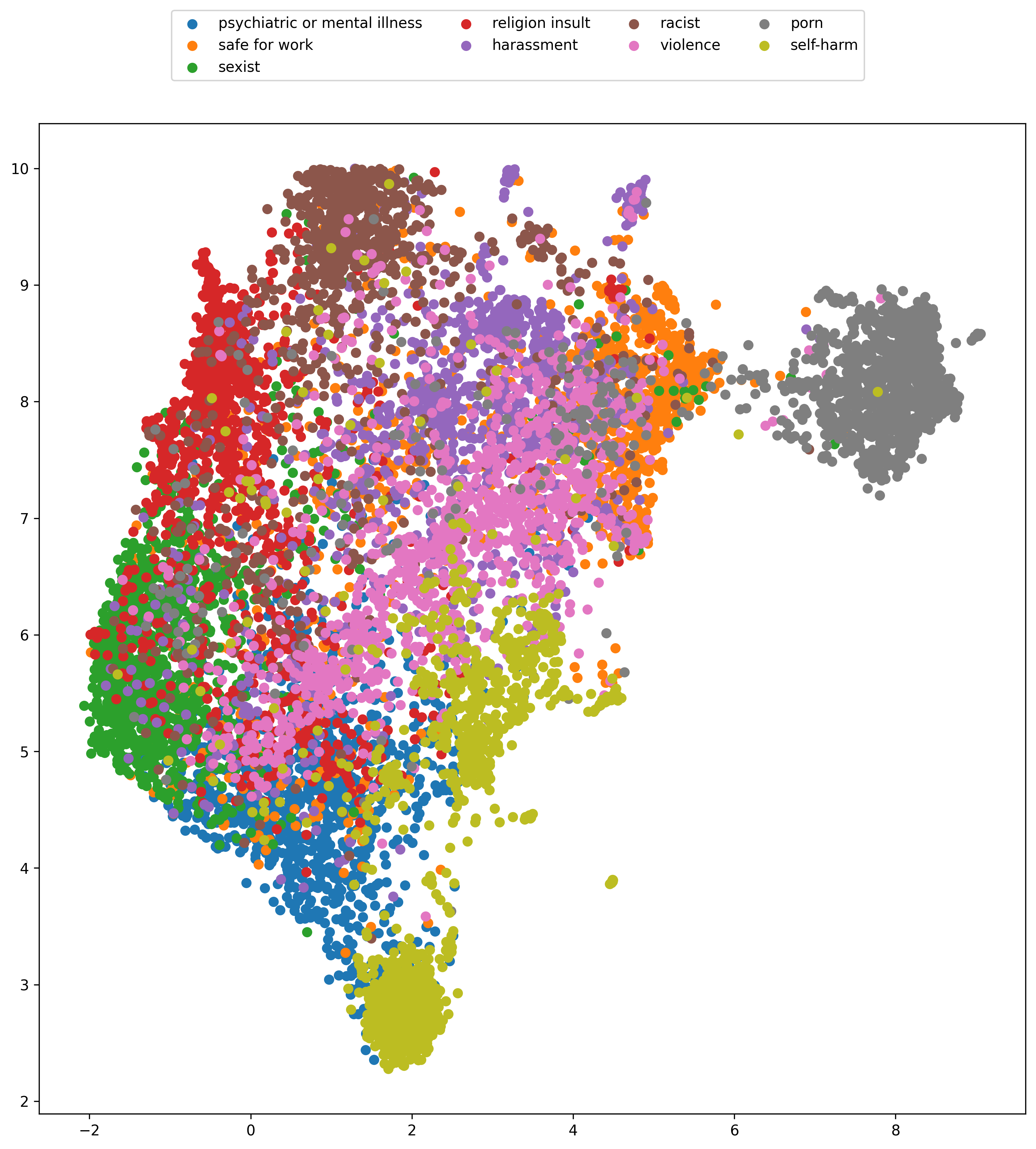}
  \caption{2D visualization using UMAP}
\end{figure}

\subsubsection*{Topic Modeling}

For topic modeling, we use Term Frequency-Inverse Document Frequency (TFIDF) as the vectorization method with parameters \texttt{max\_df=0.95} and \texttt{min\_df=1}. Additionally, we utilize stop words from the Malaya Toolkit \cite{Malaya} to filter out common words that may not contribute meaningfully to the topics. For the decomposition step, we employ Latent Dirichlet Allocation (LDA) from the Scikit-learn library \cite{scikit-learn}, and generate the top 10 topics.

\begin{table}[h]
  \centering
  \begin{tabularx}{\textwidth}{|X|X|}
    \hline
    \textbf{Label}                & \textbf{Topics}                                                                                  \\
    \hline
    Self-Harm                     & "bunuh", "mati", "life", "suicidal", "tidur", "people", "hidup", "anymore", "suicide", "die"     \\
    Harassment                    & "sial", "babi", "jialat", "mampus", "pukimak", "perangai", "bodoh", "gila", "anak", "sampah"     \\
    Porn                          & "lucah", "seks", "cerita", "melayu", "filem", "xxx", "sex", "fuck", "wanita", "kanak"            \\
    Religion Insult               & "islam", "agama", "jelek", "people", "kristen", "muslim", "solat", "religious", "anak", "quran"  \\
    Sexist                        & "lelaki", "women", "husband", "perempuan", "anak", "wife", "suami", "men", "isteri", "woman"     \\
    Psychiatric or Mental illness & "anak", "mother", "hati", "husband", "family", "heart", "people", "suami", "life", "child"       \\
    Safe For Work                 & "good", "beautiful", "cute", "love", "people", "happy", "watch", "friend", "malaysia", "anak"    \\
    Violence                      & "rumah", "anak", "kereta", "polis", "kes", "mati", "masuk", "adik", "jalan", "mangsa"            \\
    Racist                        & "melayu", "people", "malay", "chinese", "malays", "malaysia", "cina", "racist", "bangsa", "race" \\
    \hline
  \end{tabularx}
  \caption{Top 10 unigram topics using TFIDF-LDA for each label.}
  \label{table:unigram-topics}
\end{table}

\newpage

\begin{table}[h]
  \centering
  \begin{tabularx}{\textwidth}{|X|X|}
    \hline
    \textbf{Label}                & \textbf{Topics}                                                                                                                                                      \\
    \hline
    Self-Harm                     & "putus asa", "suicidal thoughts", "terjun bangunan", "suicidal thought", "cekik mati", "makan pil", "binge purge", "asa hidup", "lompat bangunan", "bunuh suicidal"  \\
    Harassment                    & "gila babi", "jialat jialat", "padan muka", "sial perangai", "damn jialat", "babi sial", "perangai sial", "pukimak sial", "pergi mampus", "cam sial"                 \\
    Porn                          & "cerita lucah", "kanak kanak", "hubungan seks", "cerita seks", "lucah seks", "si rambut", "lucah filem", "kanak perempuan", "filem lucah", "lucah melayu"            \\
    Religion Insult               & "agama islam", "umat islam", "islam jelek", "pakai tudung", "ngata ngatain", "agama kristen", "prophet muhammad", "religious teacher", "hina agama", "mengolok olok" \\
    Sexist                        & "anak anak", "kerja rumah", "anak perempuan", "household chores", "anak lelaki", "care children", "men women", "taking care", "jaga anak", "lelaki perempuan"        \\
    Psychiatric or Mental illness & "anak anak", "air mata", "adik beradik", "kawan kawan", "terima kasih", "kasih sayang", "taking care", "younger sibling", "sakit hati", "sorang sorang"              \\
    Safe For Work                 & "terima kasih", "laugh loud", "pakatan harapan", "originally posted", "good luck", "happy birthday", "pompom pompom", "ayuh malaysia", "anak anak", "adik beradik"   \\
    Violence                      & "anak anak", "adik beradik", "meninggal dunia", "kejadian berlaku", "report polis", "kanak kanak", "pakatan harapan", "death penalty", "masuk bilik", "adik adik"    \\
    Racist                        & "ang moh", "ketuanan melayu", "ah beng", "white people", "chinese people", "cina india", "malay people", "bangsa melayu", "mat salleh", "melayu islam"               \\
    \hline
  \end{tabularx}
  \caption{Top 10 bigram topics using TFIDF-LDA for each label.}
  \label{table:bigram-topics}
\end{table}

By examining the top words for each topic, we can gain an understanding of the common themes and specific vocabulary associated with each label. For example, words related to self-harm include "putus asa" (gave up), "makan pil" (pill suicide), and "lompat bangunan" (Suicide by jumping from height), while harassment-related words include strong negative expressions such as "sial" (damn) and "babi sial" (pig) that use in Malaysian context.

\subsubsection*{Wordcloud}

In addition to the tabulated results, we also created word clouds to visually represent the most prominent words for each topic. These word clouds provide an intuitive overview of the vocabulary distribution within each label category.

\newpage

\begin{figure}[h]
  \centering
  \begin{minipage}[b]{0.45\textwidth}
    \centering
    \includegraphics[width=\linewidth]{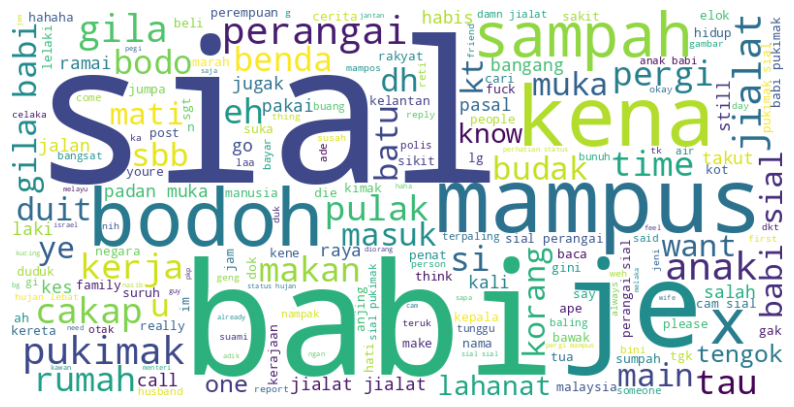}
    \captionof{figure}{Harassment Wordcloud}
  \end{minipage}
  \hfill
  \begin{minipage}[b]{0.45\textwidth}
    \centering
    \includegraphics[width=\linewidth]{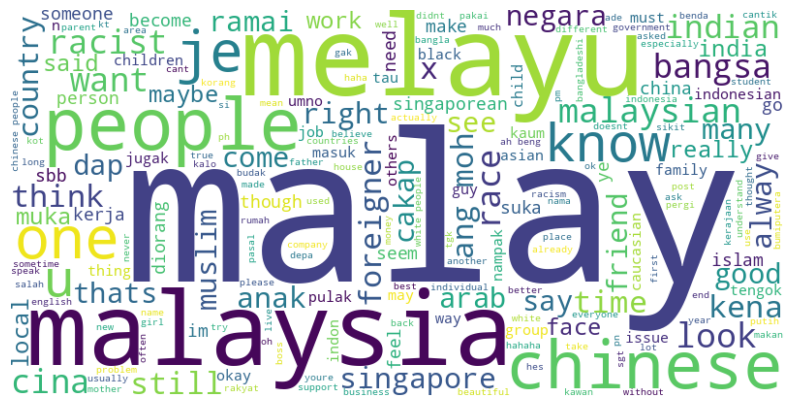}
    \captionof{figure}{Racist Wordcloud}
  \end{minipage}
\end{figure}

\begin{figure}[h]
  \centering
  \begin{minipage}[b]{0.45\textwidth}
    \centering
    \includegraphics[width=\linewidth]{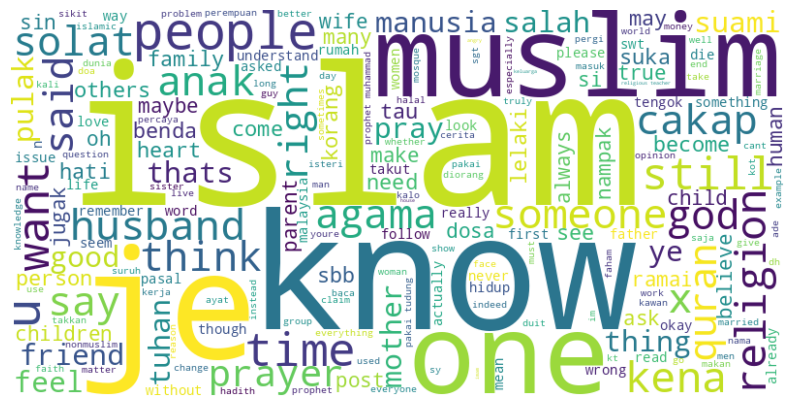}
    \captionof{figure}{Religion Insult Wordcloud}
  \end{minipage}
  \hfill
  \begin{minipage}[b]{0.45\textwidth}
    \centering
    \includegraphics[width=\linewidth]{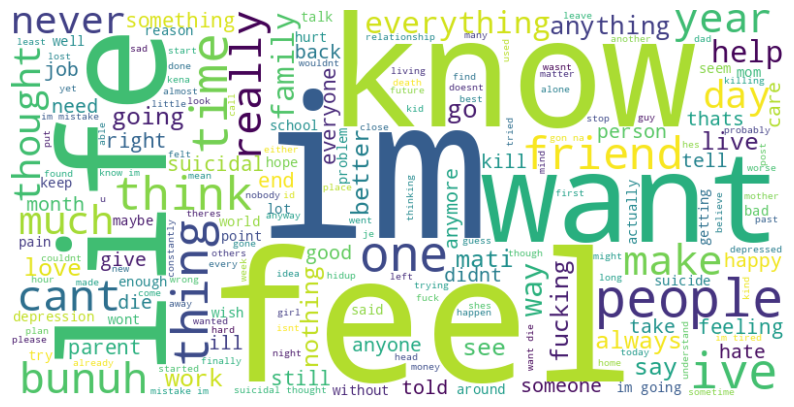}
    \captionof{figure}{Self-Harm Wordcloud}
  \end{minipage}
\end{figure}

\begin{figure}[h]
  \centering
  \begin{minipage}[b]{0.45\textwidth}
    \centering
    \includegraphics[width=\linewidth]{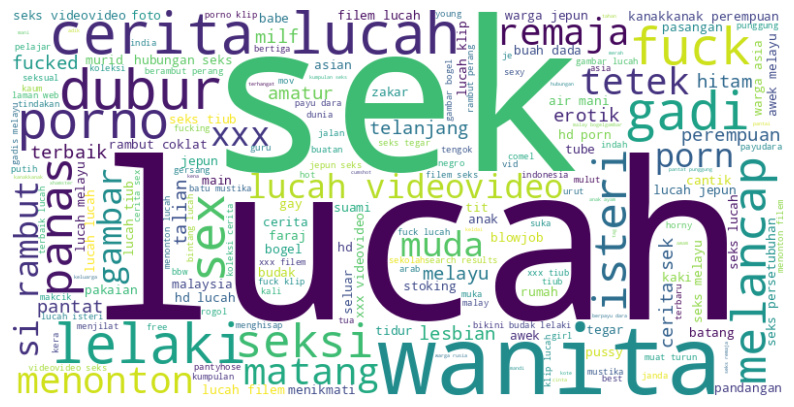}
    \captionof{figure}{Porn Wordcloud}
  \end{minipage}
  \hfill
  \begin{minipage}[b]{0.45\textwidth}
    \centering
    \includegraphics[width=\linewidth]{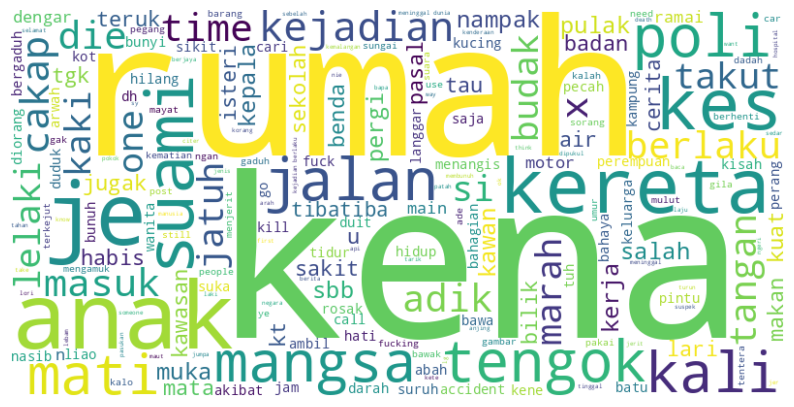}
    \captionof{figure}{Violence Wordcloud}
  \end{minipage}
\end{figure}

\begin{figure}[h]
  \centering
  \begin{minipage}[b]{0.45\textwidth}
    \centering
    \includegraphics[width=\linewidth]{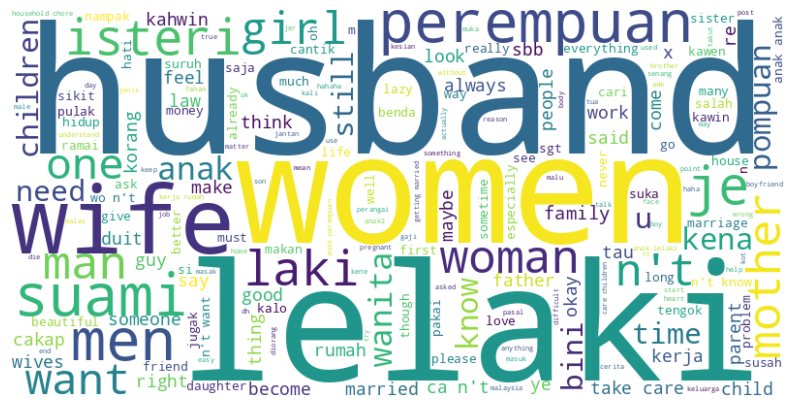}
    \captionof{figure}{Sexist Wordcloud}
  \end{minipage}
  \hfill
  \begin{minipage}[b]{0.45\textwidth}
    \centering
    \includegraphics[width=\linewidth]{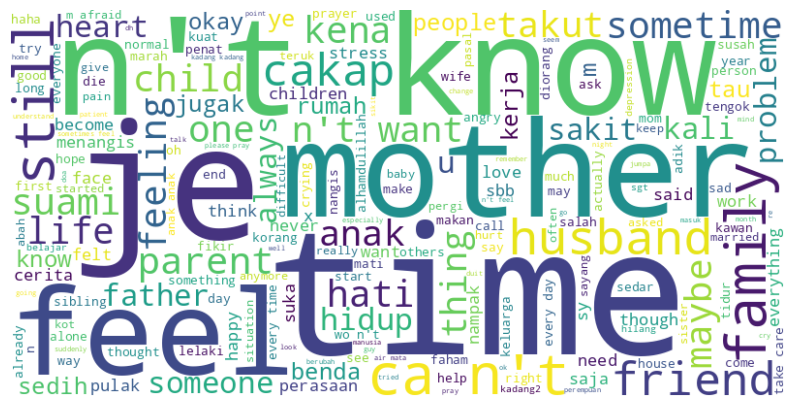}
    \captionof{figure}{Psychiatric or Mental Illness Wordcloud}
  \end{minipage}
\end{figure}

\newpage

These word clouds further emphasize the differences in language and focus across the various labels. For instance, the "Porn" word cloud is dominated by words like "seks" (sex) and "lucah" (porn), while the "Racist" word cloud includes terms such as "melayu" (Malay), "cina" (Chinese), and "DAP" (Democratic Action Party).

\section{Conclusion}

In conclusion, our work has presented a novel approach to moderating harmful content in AI safety through the introduction of a Safe For Work (SFW) classifier. By harnessing the power of large language models and data mining techniques, we have constructed a valuable dataset annotated with harmful topics and the first SFW classifier for Malaysian texts. This pioneering research fills a significant gap in the field, as no prior studies have explored this specific approach. Encouraged by our promising results, we leave for future work the refinement of the classifier to better distinguish among varying levels of harmful content and different types of harmful topics. Given the vast amount of user-generated content online, we believe this work represents a significant step forward in AI safety and content moderation.

\section{Acknowledgement}

We would like to express our gratitude to Mesolitica for providing us with the resources to train our models. Their support has played a crucial role in the success of our research, enabling us to leverage advanced technologies and computational resources.

We extend our thanks to the wider research community for their valuable insights and collaborative discussions, which have greatly influenced our work. This paper reflects the collective efforts and contributions from both NVIDIA Inception and the broader research community.

\bibliography{neurips_2023}{}

\begin{thebibliography}{10}

\bibitem{markov2023holistic}
Todor Markov, Chong Zhang, Sandhini Agarwal, Tyna Eloundou, Teddy Lee, Steven Adler, Angela Jiang, and Lilian Weng.
\newblock A holistic approach to undesired content detection in the real world, 2023.

\bibitem{vaswani2023attentionneed}
Ashish Vaswani, Noam Shazeer, Niki Parmar, Jakob Uszkoreit, Llion Jones, Aidan~N. Gomez, Lukasz Kaiser, and Illia Polosukhin.
\newblock Attention is all you need, 2023.

\bibitem{qiu2024facilitatingpornographictextdetection}
Huachuan Qiu, Shuai Zhang, Hongliang He, Anqi Li, and Zhenzhong Lan.
\newblock Facilitating pornographic text detection for open-domain dialogue systems via knowledge distillation of large language models, 2024.

\bibitem{LabelStudio}
Maxim Tkachenko, Mikhail Malyuk, Andrey Holmanyuk, and Nikolai Liubimov.
\newblock Label studio: Data labeling software, 2020-2022.
\newblock Open source software available from https://github.com/heartexlabs/label-studio.

\bibitem{zolkepli2024mallam}
Husein Zolkepli, Aisyah Razak, Kamarul Adha, and Ariff Nazhan.
\newblock Mallam -- malaysia large language model, 2024.

\bibitem{behnamghader2024llm2veclargelanguagemodels}
Parishad BehnamGhader, Vaibhav Adlakha, Marius Mosbach, Dzmitry Bahdanau, Nicolas Chapados, and Siva Reddy.
\newblock Llm2vec: Large language models are secretly powerful text encoders, 2024.

\bibitem{2018arXivUMAP}
L.~{McInnes}, J.~{Healy}, and J.~{Melville}.
\newblock {UMAP: Uniform Manifold Approximation and Projection for Dimension Reduction}.
\newblock {\em ArXiv e-prints}, February 2018.

\bibitem{mcinnes2018umap-software}
Leland McInnes, John Healy, Nathaniel Saul, and Lukas Grossberger.
\newblock Umap: Uniform manifold approximation and projection.
\newblock {\em The Journal of Open Source Software}, 3(29):861, 2018.

\bibitem{Malaya}
Zolkepli Husein.
\newblock Malaya.
\newblock \url{https://github.com/huseinzol05/malaya}, 2018.

\bibitem{scikit-learn}
F.~Pedregosa, G.~Varoquaux, A.~Gramfort, V.~Michel, B.~Thirion, O.~Grisel, M.~Blondel, P.~Prettenhofer, R.~Weiss, V.~Dubourg, J.~Vanderplas, A.~Passos, D.~Cournapeau, M.~Brucher, M.~Perrot, and E.~Duchesnay.
\newblock Scikit-learn: Machine learning in {P}ython.
\newblock {\em Journal of Machine Learning Research}, 12:2825--2830, 2011.

\end{thebibliography}
\bibliographystyle{unsrt}

\end{document}